\documentclass[journal]{IEEEtran}
\usepackage{amsmath,amsfonts}
\usepackage{algorithm}
\usepackage{algpseudocode}
\usepackage{amsmath}
\usepackage{array}
\usepackage[caption=false,font=normalsize,labelfont=sf,textfont=sf]{subfig}
\usepackage{textcomp}
\usepackage{stfloats}
\usepackage{url}
\usepackage{verbatim}
\usepackage{graphicx}
\usepackage{subcaption}
\usepackage{cite}
\usepackage[hidelinks,colorlinks=false]{hyperref}
\usepackage{color}
\usepackage{booktabs}

\usepackage{multirow}
\usepackage{diagbox}
\usepackage{scalerel}
\usepackage{tikz}
\usetikzlibrary{svg.path}

\definecolor{orcidlogocol}{HTML}{A6CE39}
\tikzset{
  orcidlogo/.pic={
    \fill[orcidlogocol] svg{M256,128c0,70.7-57.3,128-128,128C57.3,256,0,198.7,0,128C0,57.3,57.3,0,128,0C198.7,0,256,57.3,256,128z};
    \fill[white] svg{M86.3,186.2H70.9V79.1h15.4v48.4V186.2z}
                 svg{M108.9,79.1h41.6c39.6,0,57,28.3,57,53.6c0,27.5-21.5,53.6-56.8,53.6h-41.8V79.1z M124.3,172.4h24.5c34.9,0,42.9-26.5,42.9-39.7c0-21.5-13.7-39.7-43.7-39.7h-23.7V172.4z}
                 svg{M88.7,56.8c0,5.5-4.5,10.1-10.1,10.1c-5.6,0-10.1-4.6-10.1-10.1c0-5.6,4.5-10.1,10.1-10.1C84.2,46.7,88.7,51.3,88.7,56.8z};
  }
}

\newcommand\orcidicon[1]{\href{https://orcid.org/#1}{\mbox{\scalerel*{
\begin{tikzpicture}[yscale=-1,transform shape]
\pic{orcidlogo};
\end{tikzpicture}
}{|}}}}

\begin{document}

\title{High-Precision 6DOF Pose Estimation via Global Phase Retrieval in Fringe Projection Profilometry for 3D Mapping}

\author{Sehoon Tak$^{\textsuperscript{\orcidicon{0009-0008-0291-4005}}}$, Keunhee Cho$^{\textsuperscript{\orcidicon{0009-0004-5862-4163}}}$,
Sangpil Kim$^{\textsuperscript{\orcidicon{0000-0002-7349-0018}}}$,
and Jae-Sang Hyun$^{\textsuperscript{\orcidicon{0000-0003-1711-8243}}}$

\thanks{Corresponding author: Sangpil Kim and Jae-Sang Hyun}
\thanks{Sehoon Tak, Keunhee Cho, and Jae-Sang Hyun are with the Department of Mechanical Engineering, Yonsei University, Seoul 03722, South Korea. (e-mail: hyun.jaesang@yonsei.ac.kr) and Sangpil Kim is with the Department of Artificial Intelligence, Korea University, Seoul (e-mail:spk7@korea.ac.kr)}}




\maketitle

\begin{abstract}
Digital fringe projection (DFP) enables micrometer-level 3D reconstruction, yet extending it to large-scale mapping remains challenging because six-degree-of-freedom pose estimation often cannot match the reconstruction’s precision. Conventional iterative closest point (ICP) registration becomes inefficient on multi-million-point clouds and typically relies on downsampling or feature-based selection, which can reduce local detail and degrade pose precision. Drift-correction methods improve long-term consistency but do not resolve sampling sensitivity in dense DFP point clouds.
We propose a high-precision pose estimation method that augments a moving DFP system with a fixed, intrinsically calibrated global projector. Using the global projector’s phase-derived pixel constraints and a PnP-style reprojection objective, the method estimates the DFP system pose in a fixed reference frame without relying on deterministic feature extraction, and we experimentally demonstrate sampling invariance under coordinate-preserving subsampling. Experiments demonstrate sub-millimeter pose accuracy against a reference with quantified uncertainty bounds, high repeatability under aggressive subsampling, robust operation on homogeneous surfaces and low-overlap views, and reduced error accumulation when used to correct ICP-based trajectories. The method extends DFP toward accurate 3D mapping in quasi-static scenarios such as inspection and metrology, with the trade-off of time-multiplexed acquisition for the additional projector measurements.

\end{abstract}


\section{Introduction}

\IEEEPARstart{T}{he} digital fringe projection (DFP) system, consisting of a digital projector and a camera, has reached micrometer-level 3D reconstruction resolution and high point density per frame \cite{ZHANG2018119}. One of the well-known methods in DFP, phase-shifting profilometry (PSP), achieves dense reconstruction by projecting phase-shifted fringe patterns and recovering an absolute phase map for each camera pixel. Owing to its measurement fidelity, DFP has been extensively employed in fields such as robotics, medical imaging, and augmented/virtual reality (AR/VR), where accurate geometry and surface detail are required \cite{son2025fringe, hyun2019evaluation}. Nevertheless, extending DFP to large-scale environments remains challenging. When a DFP sensor is repositioned to cover a wider area, the resulting multi-view point clouds must be registered into a single map, which requires accurate six-degree-of-freedom (6-DoF) pose estimation with a precision commensurate with the reconstruction itself \cite{xu2020status}.

The widely used iterative closest point (ICP) family estimates relative pose by minimizing geometric residuals between successive point clouds, and robust dense variants exist to improve convergence in many regimes. \cite{9336308, chetverikov2005robust}. In dense DFP mapping, however, ICP faces two practical constraints. First, operating directly on multi-million-point frames is expensive, so registration pipelines typically reduce the input set (random/uniform downsampling, voxel filtering, or feature-driven selection) to meet runtime constraints. Second, this reduction is not neutral: the resulting pose can become sensitive to the sampling scheme and resolution, which limits repeatability and can introduce local misalignment even when the underlying 3D measurement is high fidelity. These issues are amplified in two common DFP cases: (i) homogeneous or weakly featured surfaces, where feature-driven sampling and correspondence pruning are unreliable, and (ii) small view overlap, where correspondence-based registration becomes ill-conditioned. As a result, conventional ICP-based pipelines often trade measurement detail for tractability. Consequently, the resulting pose estimates may no longer reflect the intrinsic resolution of the reconstructed point cloud.

To reduce drift and improve global consistency, many systems introduce additional constraints. Inertial measurement units (IMUs) can provide independent motion cues and help correct accumulated error when the estimate deviates from a reference \cite{8258997}. More complex self-correction methods such as loop closure \cite{ho2007detecting} and pose-graph optimization \cite{carlone2015initialization} have been used to mitigate long-term drift in SLAM-type scenarios \cite{glocker2013real}. These approaches improve global consistency, but they do not remove the need for downsampling nor resolve the local sampling sensitivity that limits pose repeatability on high-density DFP point clouds. In particular, when the underlying registration step is unstable on featureless surfaces or small overlaps, global corrections may still converge to locally misaligned maps.

Learning-based pose estimation has also shown progress in constrained settings. DeepVO infers pose directly from monocular RGB images using a CNN/RNN architecture \cite{7989236}, and UnDeepVO extends this to stereo images using unsupervised learning \cite{8461251}. However, such models are typically trained on domain-specific datasets and are not designed to exploit phase-derived metrology constraints available in DFP. Moreover, direct comparisons at the measurement precision level of structured light systems are rarely reported for dense, real-time point-cloud mapping tasks. As a result, learning-based approaches are not currently a drop-in solution for sampling-robust, metrologically traceable pose estimation in DFP.

In this paper, we propose a high-precision pose estimation method that augments a moving DFP system with an additional fixed (global) projector. The key idea is to use the global projector's phase-derived pixel constraints as a stable reference and estimate the DFP system pose by minimizing a PnP-style reprojection objective, rather than relying on point-to-point correspondences in the reconstructed clouds. The method incorporates a batched sampling protocol and a consensus-based rejection strategy to obtain stable estimates from small point subsets, and it is characterized to be robust to coordinate-preserving subsampling (random, uniform, and voxel-nearest schemes) across a wide range of point counts. The experimental evaluation focuses on quasi-static acquisitions under time-multiplexed projection, and estimator-only throughput is reported separately.

We can formalize the problem statement as follows; given a reconstructed local point set $\mathbf{X}_C$ (in the camera frame) and the global projector phase-derived pixel constraints
$\{\mathbf{u}_i^g\}$ observed at the same camera pose, estimate the absolute 6-DoF pose in the fixed global projector reference frame. 
We solve for this pose by minimizing a PnP-style reprojection objective with coordinate-preserving batched subsampling and consensus filtering, and refine using a geometry-consistency regularizer (Sec.~\ref{sec:model}).

To meet metrological and reproducibility requirements, we report calibration residuals and reconstruction repeatability, and we provide a practical sensitivity analysis linking calibration/measurement uncertainty to pose drift. We then validate (i) estimator repeatability and sampling robustness on fixed reconstructions, (ii) robustness under reduced overlap, (iii) featureless surface registration where ICP is unreliable, and (iv) reduced error accumulation when the proposed estimator is used to correct ICP-based trajectories, including statistical significance testing. While the method substantially reduces sampling-induced variability compared to conventional ICP pipelines, small biases can remain and are reported through the metrological and sensitivity analyses.

The rest of the paper is organized as follows. Section 2 reviews PSP phase retrieval and the projection model used to form pixel-level constraints. Section 3 describes the system configuration, batching/sampling protocol, and optimization objectives. Section 4 provides metrological characterization and estimator sensitivity/robustness studies, including calibration statistics, reconstruction repeatability, propagation analysis, and sampling/overlap characterization. Section 5 presents application experiments on featureless surface registration and error accumulation reduction in ICP trajectories, followed by statistical testing. Section 6 concludes with limitations and practical considerations.

\section{Principles}
\label{sec:principles}
\subsection{Absolute Phase Retrieval}
\label{subsec:absolute}
In phase-shifting profilometry (PSP) and related structured-light (SL) systems, a 3D reconstruction is typically obtained by (i) phase retrieval, (ii) phase unwrapping, and (iii) conversion from camera--projector correspondences to 3D coordinates using calibration data. Phase-shifting methods are widely used for phase retrieval due to their efficiency and robustness to intensity noise.

The minimum number of phase-shifted fringe images required for phase retrieval is three. While three-step methods enable fast reconstruction, higher-step methods are often preferred for improved robustness and phase resolution. In this work, we use an $N$-step phase-shifting algorithm~\cite{ZUO201823}, where $N$ phase-shifted patterns are projected for a single 3D measurement:
\begin{equation}
\label{n_step}
I_{n}(u,v)=I^{'}(u,v)+I^{''}(u,v)\cos[\phi(u,v)-2\pi n/N],
\end{equation}
where $n$ is the phase-shift index ($n=0,1,\ldots,N-1$), $(u,v)$ denotes camera pixel coordinates, $I^{'}$ is the average intensity, $I^{''}$ is the modulation, and $2\pi n/N$ is the known phase shift. The wrapped phase $\phi$ at each pixel is obtained as
\begin{equation}
\label{phase_extract}
\phi(u, v) = \operatorname{atan2}\!\left(
\begin{aligned}
&\sum^{N-1}_{n=0} I_n(u, v)\sin(2\pi n/N),\\
&\sum^{N-1}_{n=0} I_n(u, v)\cos(2\pi n/N)
\end{aligned}
\right).
\end{equation}
We implement Eq.~\ref{phase_extract} using $\operatorname{atan2}(\cdot)$ to ensure correct quadrant selection.

The wrapped phase lies in $(-\pi,\pi]$ and contains $2\pi$ discontinuities due to the inverse tangent function. To obtain continuous and uniquely-defined projector coordinates, the wrapped phase is converted to an unwrapped (absolute) phase map, denoted $\Phi$. In this work we use temporal Gray-code phase unwrapping~\cite{ZUO201684}. By projecting binary-coded patterns in addition to the phase-shifted fringes, each pixel is assigned a unique fringe index (fringe order) that resolves the $2\pi$ ambiguity:
\begin{equation}
\label{fringe_order}
\Phi(u^{c},v^{c})=\phi(u^{c},v^{c})+2\pi\,k(u^{c},v^{c}),
\end{equation}
where $k$ is the fringe order.

\subsection{Pinhole Projection Model and Phase-to-Pixel Mapping}
\label{subsec:pinhole}
Given a calibrated camera--projector pair, absolute phase maps can be converted to projector pixel coordinates. For a horizontal/vertical phase encoding, the mapping is proportional:
\begin{equation}
\begin{aligned}
\label{phase_to_coord}
u^p &= \Phi_{\text{h}}(u^c, v^c)\frac{T_H}{2\pi}, \\
v^p &= \Phi_{\text{v}}(u^c, v^c)\frac{T_V}{2\pi}
\end{aligned}
\end{equation}
where $(u^c,v^c)$ are camera pixel coordinates, $(u^p,v^p)$ are projector pixel coordinates, and $T_H$, $T_V$ are scale factors determined by the projected fringe periods (i.e., the phase-to-pixel conversion constants for the pattern design).

Projector image formation is modeled using the standard pinhole projection model. Let $\tilde{\mathbf{u}}^{p}=[\tilde{u}^{p},\tilde{v}^{p},\tilde{w}^{p}]^\top$ denote homogeneous projector image coordinates and let $\pi(\cdot)$ denote dehomogenization (perspective division). The projector projection equation is
\begin{equation}
\label{eq_pinhole}
\tilde{\mathbf{u}}^{p} =
\mathbf{K}^{p}\left[\mathbf{R}\,|\,\mathbf{t}\right]
\begin{bmatrix}\mathbf{X}\\1\end{bmatrix},
\qquad
\mathbf{u}^{p} =
\pi(\tilde{\mathbf{u}}^{p})=
\begin{bmatrix}
\tilde{u}^{p}/\tilde{w}^{p}\\
\tilde{v}^{p}/\tilde{w}^{p}
\end{bmatrix},
\end{equation}
where $\mathbf{u}^{p}=[u^{p},v^{p}]^\top$ are projector pixel coordinates, $\mathbf{K}^{p}$ is the projector intrinsic matrix, and $\left[\mathbf{R}\,|\,\mathbf{t}\right]$ maps 3D points from the chosen reference frame to the projector coordinate frame.

In many SL pipelines, reconstructed 3D points are expressed in the camera coordinate frame by default. This is because triangulation uses the calibrated camera and projector models. In this paper, we keep that convention internally where it is natural (e.g., for local reconstruction and for ICP), but we report all final poses and errors in a single global reference frame (defined in Sec.\ref{subsec:configuration}) to ensure consistent evaluation and to avoid frame ambiguity in figures and tables.

\section{The Optimization Model}
\label{sec:model}
Pose estimation is formulated as an extrinsic calibration problem between a moving local reconstruction system and a fixed, intrinsically calibrated global projector. Similar to PnP and related reprojection-based estimators\cite{9549863}, we use a reprojection error objective defined in the global projector image domain. However, unlike conventional PnP or ICP pipelines that rely on dense correspondences or deterministic keypoint selection, the proposed method is designed to remain stable under aggressive sub-sampling of high-density structured-light reconstructions by (i) filtering measurement-unsafe pixels and (ii) estimating pose from multiple small, independent batches and rejecting inconsistent batches.

This section defines the system configuration (Sec.~\ref{subsec:configuration}), the batching and sampling protocol (Sec.~\ref{subsec:batching}), optimization objective and refinement strategy (Sec.~\ref{subsec:optimization}), and computational profile (Sec.~\ref{subsec:implementation}). Experimental justification of sampling invariance, overlap robustness, and hyperparameter choices is reported later in Sec.~\ref{sec:metrology}.

\begin{figure}[!h]
\centering
\includegraphics[width=\columnwidth, trim = 0 180 0 180, clip]{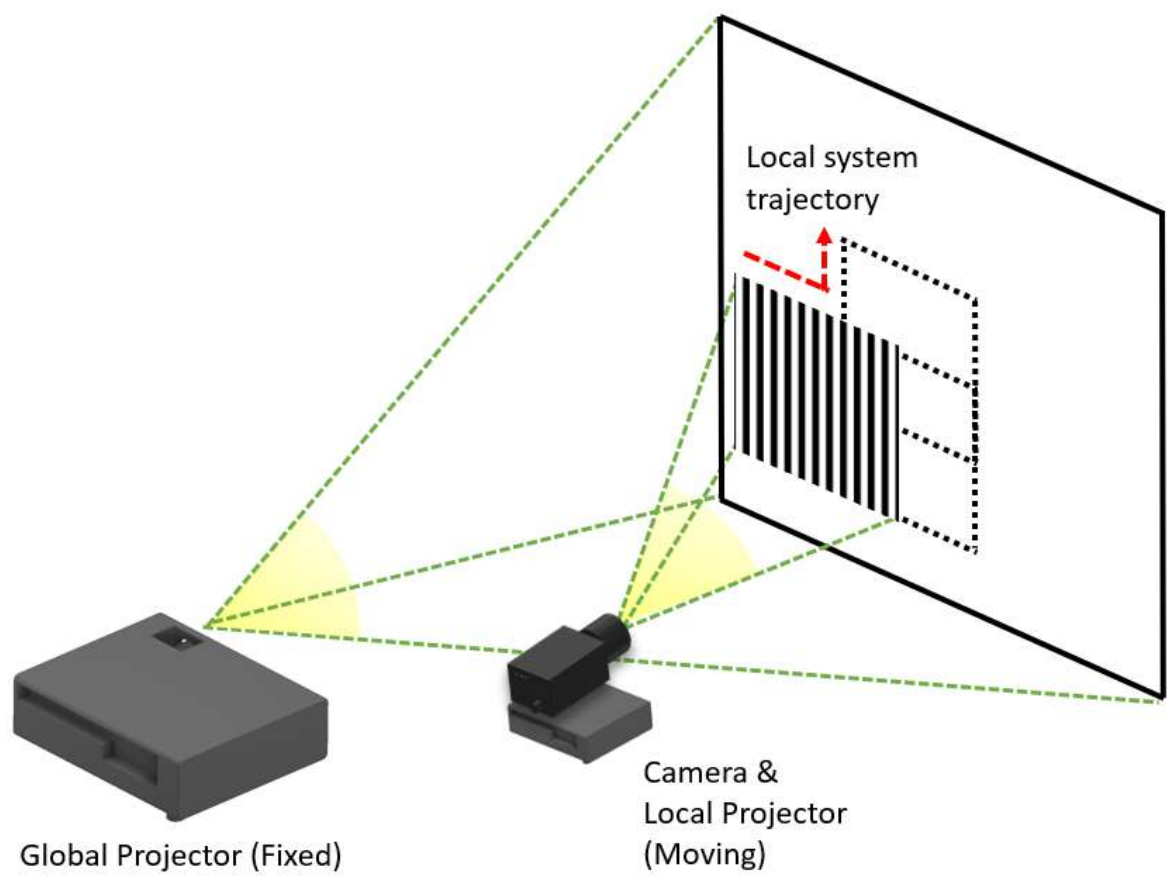}
\caption{Illustration of system setup.}
\label{setup_img}
\end{figure}

\subsection{System Configuration and Measurement-Validity Filtering}
\label{subsec:configuration}
First, we fix the coordinate frame definitions. Let $\{C\}$ denote the local camera frame, $\{P_g\}$ the fixed global projector frame (used as the global reporting frame), and $\{P_\ell\}$ the local projector frame used for structured-light reconstruction. For each acquisition, the local DFP subsystem produces a reconstructed point cloud $\mathbf{X}_C=\{ \mathbf{x}_i \in \mathbb{R}^3 \}$ expressed in $\{C\}$, together with a global-projector pixel coordinate $(u_i^{g}, v_i^{g})$ obtained from the global projector phase map at the same camera pixel. The global projector intrinsics $\mathbf{K}^{g}$ are assumed known from calibration (Sec.~\ref{subsec:calibration}). The goal of the estimator is to recover the rigid transform ${}^{P_g}\mathbf{T}_{C}=[\mathbf{R}\,|\,\mathbf{t}]$ that maps camera-frame points into the global projector frame for reprojection.

The estimator assumes no explicit surface priors (e.g., planarity, texture features) and instead relies on rejecting measurements that are likely to violate the phase-to-pixel mapping or triangulation assumptions. Concretely, we remove pixels that are likely to carry phase ambiguity or reconstruction bias, including:
(i) saturated/underexposed pixels identified from the phase-shifting intensity stack,
(ii) pixels near Gray-code transition regions (high fringe-order risk),
(iii) pixels affected by occlusion/self-shadowing where modulation is insufficient,
and (iv) pixels close to fringe edges where phase gradient is unstable.
These conditions can be detected during reconstruction without requiring global pose and therefore can be applied online as part of the measurement pipeline. The resulting validity mask is applied before sampling so that sampled points are drawn from a set with approximately uniform measurement quality.

In general, only the camera pixels simultaneously observing valid local reconstruction and valid global projector phase contribute constraints. This overlap may be small when the local subsystem approaches the edge of the global projector field of view. Robustness to small overlap is characterized experimentally in Sec.~\ref{subsubsec:overlap}; empirically, the method remains stable provided that a minimum number of valid points are available and are not degenerate with respect to the 6-DoF pose.

\subsection{Batching and Sampling Protocol}
\label{subsec:batching}
A single local structured-light reconstruction typically contains on the order of $10^6$--$10^7$ points, which is too large for dense correspondence methods to process within a strict time budget. However, reprojection-based pose estimation is overdetermined: in the absence of degeneracy, a small subset of 3D--2D constraints is sufficient to estimate a 6-DoF transform. The proposed estimator therefore operates on many small \emph{batches} of sampled points rather than on the full cloud.

Let $\mathcal{V}$ denote the set of valid pixels after measurement-validity filtering (Sec.~\ref{subsec:configuration}). We draw $B$ batches per frame. Each batch samples $n$ points, where the $n$ camera pixels are chosen from $\mathcal{V}$ using a grid-based selection process in the image frame that does not contain invalid points. Each sampled camera pixel yields:
(i) a 3D point $\mathbf{x}_i$ expressed in $\{C\}$ from local reconstruction, and
(ii) a global projector pixel $(u_i^g, v_i^g)$ from global phase decoding.
Each batch thus forms a set of $n$ projector reprojection constraints under the unknown transform ${}^{P_g}\mathbf{T}_{C}$.

For a batch of $n$ constraints, let $\mathbf{U}_g\in\mathbb{R}^{2\times n}$ collect the corresponding global-projector pixels in 2D form,
$\mathbf{U}_g=[\mathbf{u}^g_1,\dots,\mathbf{u}^g_n]$ with $\mathbf{u}^g_i=[u_i^g,v_i^g]^\top$ (inhomogeneous 2D pixels).
Let $\mathbf{X}^{(b)}_C\in\mathbb{R}^{4\times n}$ stack the sampled camera-frame points in homogeneous coordinates.
We define the batch reprojection residual as
\begin{equation}
\label{eq_batch_residual}
\begin{split}
\hat{\mathbf{U}}_g(\mathbf{R},\mathbf{t}) &= \pi\!\left(\mathbf{K}^g,[\mathbf{R}|\mathbf{t}]\ ,\mathbf{X}^{(b)}_C\right),\\
\mathbf{e}_\text{reproj}(\mathbf{R},\mathbf{t}) &= \mathbf{U}_g - \hat{\mathbf{U}}_g(\mathbf{R},\mathbf{t}).
\end{split}
\end{equation}
where $\pi(\cdot)$ denotes the standard pinhole projection returning inhomogeneous 2D pixels after division by the third homogeneous component. 

The key design choice is to solve for pose from many independent small batches and later reconcile their results. This structure enables parallel implementation and provides a natural mechanism for outlier rejection by identifying batches whose optimized poses are inconsistent with the majority.

\subsection{Camera Pose Optimization}
\label{subsec:optimization}
We optimize the pose in $SE(3)$ using a 6-DoF parameter vector
$\boldsymbol{\theta}=[\varphi,\vartheta,\psi,t_x,t_y,t_z]^\top$, where $(\varphi,\vartheta,\psi)$ are roll--pitch--yaw (Euler) angles and $\mathbf{t}=[t_x,t_y,t_z]^\top$ is translation in meters. To improve numerical stability in early iterations, the translation can optionally be reparameterized with a bounded mapping by optimizing an unconstrained variable $\tilde{\mathbf{t}}$ and applying $\mathbf{t}=t_\text{max}\tanh(\tilde{\mathbf{t}})$, where $t_\text{max} = 2$ denotes a conservative translation range. 

For a batch of sampled constraints $\{(\mathbf{x}_i,(u_i^g,v_i^g))\}_{i=1}^{n}$, we minimize the global projector reprojection residual. Let $\pi(\mathbf{K}^g,{}^{P_g}\mathbf{T}_{C},\mathbf{x}_i)$ denote the projected pixel coordinate under the global projector intrinsics $\mathbf{K}^g$. The stage-1 objective is
\begin{equation}
\label{eq_loss1}
\begin{split}
\boldsymbol{\theta}^{(1)} &= \arg\min_{\boldsymbol{\theta}\in\mathbb{R}^6} f_1(\boldsymbol{\theta}),\\
f_1(\boldsymbol{\theta}) &= \frac{1}{n}\sum_{i=1}^{n}
\left\|
\mathbf{u}_i^g
-
\pi\!\left(\mathbf{K}^g,\ {}^{P_g}\mathbf{T}_{C}(\boldsymbol{\theta})\ ,{\mathbf{x}}_i\right)
\right\|_2^2,
\end{split}
\end{equation}
Here $f_1$ is the mean-squared reprojection error (MSE) in global-projector pixel units, averaged over the $n$ points in the batch. The parameter vector $\boldsymbol{\theta}\in\mathbb{R}^6$ induces a rigid transform ${}^{P_g}\mathbf{T}_{C}\in SE(3)$ via parameterization. When quoting thresholds or reporting convergence diagnostics, we use the corresponding RMS reprojection error in pixels. 

We evaluated two solvers: (i) Levenberg--Marquardt (LM)~\cite{Levenberg1944} and (ii) Adam~\cite{kingma2014adam}. LM was implemented in a nonlinear least-squares framework, while Adam was implemented to support batched parallel optimization on the GPU. Under the proposed batching/consensus protocol, Adam provided higher throughput for many parallel hypotheses and was empirically more tolerant to imperfect initialization and noisy gradients from small batches, enabling reliable consensus filtering across parallel runs. In our experiments, the LM solver showed higher sensitivity to initialization at the required hypothesis scale, whereas Adam provided more stable convergence under the proposed batching protocol. In translation-dominated experiments like Sec.~\ref{subsubsec:precision}, LM failed to converge to valid pose ranges in 97.1 \% of batches, compared to that of Adam given in Table~\ref{tab:gt_timing}.

Stage-1 optimization is performed over $B$ independent batches in parallel, yielding a set of candidate poses $\{\boldsymbol{\theta}^{(1)}_b\}_{b=1}^{B}$. Because some batches may be ill-conditioned or contain residual decoding errors, we reject inconsistent batches using a consensus rule: poses whose deviation from the batch-median exceeds one standard deviation in translation and rotation are discarded, and the remaining poses are aggregated to obtain a robust estimate. This mechanism plays the role of a correspondence-free outlier filter analogous to RANSAC, but operating on pose hypotheses produced by independent sub-samples.

Reprojection error alone may admit multiple local minima, particularly when the sampled constraints are spatially clustered or when residual decoding errors remain. We therefore add a weak refinement term that encourages consistency between (i) the local reconstruction and (ii) a reconstruction implied by the global projector phase under the current pose estimate. Specifically, using the estimated transform ${}^{P_g}\mathbf{T}_{C}(\boldsymbol{\theta})$ and one of the global projector phase maps, we compute a secondary point set $\mathbf{X}^{g, (b)}_C(\boldsymbol{\theta})$ expressed in the camera frame by utilizing the inferred global projector geometry through triangulation. The stage-2 objective is
\begin{equation}
\label{eq_loss2}
\begin{aligned}
f_2(\boldsymbol{\theta}) &= f_1(\boldsymbol{\theta}) + \lambda\, f_X(\boldsymbol{\theta}), \\
f_X(\boldsymbol{\theta}) &= \frac{1}{n}\left\lVert \mathbf{X}^{(b)}_C - \mathbf{X}^{g, (b)}_C(\boldsymbol{\theta}) \right\rVert_F^2 .
\end{aligned}
\end{equation}
The second term $f_X$ is expressed in world units and is converted to pixel scale using $\lambda$; the reported reprojection metrics remain in pixels, while pose errors are evaluated in $SE(3)$ using translation error in millimeters and rotation error on $SO(3)$ as defined in Sec.~\ref{subsec:metrics}. $\lambda = 1e-2$ is a unit-balancing weight, derived from local linearization at nominal depth. In our implementation, this term is used as a secondary stabilizer: the dominant constraint remains the global projector reprojection consistency, and $\lambda$ is set such that the refinement does not override valid reprojection evidence. We observed that without a good initial estimate, the stage-2 optimizer on its own may not reliably converge to a valid pose.

Each batch optimization uses a maximum iteration count $T_{\max}=2000$ and terminates early if the relative decrease in $f_1$ (or $f_2$ in stage 2) falls below a threshold for $T_{\text{patience}}=10$ consecutive steps. The Adam learning rate is scheduled using cosine decay with $\alpha_0=10^{-2}$ and a 50-iter warmup. Optimization defaults used throughout the experiment are provided in Sec.~\ref{subsubsec:hyperparameters}. We report typical iteration counts and runtime distributions in Sec.~\ref{subsec:setup}.

Because the estimator explicitly links global projector pixel coordinates to 3D geometry through calibrated intrinsics and phase decoding, systematic biases in phase or depth can shift the recovered \emph{absolute} pose in the global frame. This differs from relative-only methods such as ICP, where a uniform additive depth bias may partially cancel in a relative transform. For this reason, we separately quantify reconstruction accuracy (Sec.~\ref{subsec:repeatability}) and propagate calibration/measurement uncertainty to pose drift via Monte-Carlo sensitivity analysis\cite{jcgm101_2008} (Sec.~\ref{subsec:propagation}), and we interpret GT-based results accordingly (Sec.~\ref{subsec:setup}).

\subsection{Implementation and Computational Profile}
\label{subsec:implementation}
The estimator consists of three computational stages executed per measurement frame: (i) 3D reconstruction, measurement-validity masking and sampling, (ii) batch-parallel pose optimization, (iii) cross-batch consensus filtering and aggregation. All batches are independent during optimization and therefore are naturally parallelizable across GPU threads or CPU cores. Consensus filtering operates on the resulting $B$ pose hypotheses and has negligible cost compared to the batch optimizations.

Let $B$ be the number of batches, $n$ the number of points per batch, and $T$ the number of optimizer iterations. The dominant per-frame compute cost is proportional to
\begin{equation}
\label{eq_complexity}
\mathcal{O}\!\left(B \cdot T \cdot n\right)
\end{equation}
projection operations and residual evaluations, plus an $\mathcal{O}(B)$ consensus step. Importantly, the complexity is independent of the total reconstructed point count $N_\text{cloud}$ except through sampling and data access. This decouples estimator compute time from reconstruction density, allowing the method to exploit high-resolution reconstructions without requiring dense correspondence computations.

Although the optimization cost scales with $B,T,n$, the runtime can be dominated by memory access when the reconstructed point cloud contains millions of points. In particular, transferring the full reconstructed cloud (or dense auxiliary maps) between CPU and GPU each frame can become a bottleneck. In our implementation, we therefore treat sampling as a first-class design choice: points are sampled from the validity mask prior to optimization, and only the sampled subsets are materialized for the optimizer.

To avoid conflating estimator compute with structured-light acquisition, we report two timing measures:
(1) \emph{estimator-only time}, measured for the pose solver given reconstructed $\mathbf{X}_C$ and global projector phase maps; and
(2) \emph{update-period budget under the experimental schedule}, obtained from the time-multiplexed projection/capture sequence length and exposure/trigger settings described in Sec.~\ref{subsec:setup}.
Timing statistics are reported as median over 120 frames. We do not claim an end-to-end real-time implementation; the timing budget is reported to quantify the acquisition overhead and to show that the estimator is not compute-limited under the tested settings.

For each batch, reprojection residual decreases rapidly during the first 80 iterations and then exhibits diminishing returns. The default iteration budgets and early-stopping thresholds are fixed across experiments and provided in Sec.~\ref{subsubsec:hyperparameters}. 

When run in isolation, estimator throughput is primarily controlled by $(B,n,T)$, the stopping criteria, and hardware parallelism. In a full system, the achievable pose update additionally depends on the structured-light sequence length and the time-multiplexing between local and global projectors. The proposed method reduces the need for costly deterministic sampling or feature extraction on dense point clouds, but it introduces additional acquisition overhead\cite{VANDERJEUGHT201618} due to global-projector constraints; this trade-off is quantified in Sec.~\ref{subsec:setup} and summarized as a limitation in Sec.~\ref{sec:conclusion}. Accordingly, the experimental evaluation emphasizes estimator accuracy / robustness and acquisition-limited timing budgets rather than demonstrating maximum end-to-end throughput.

\section{Metrological Characterization and Sensitivity}
\label{sec:metrology}
\subsection{Metrics and Reporting Conventions}
\label{subsec:metrics}
This section defines the error metrics and reporting conventions used throughout the rest of the paper. Unless otherwise noted, all poses are reported in the fixed global projector frame $\{P_g\}$, and all scalar error metrics are computed from that representation.

A pose is represented as a rigid transform ${}^{P_g}\mathbf{T}_{C}=[\mathbf{R}\,|\,\mathbf{t}]\in SE(3)$ mapping 3D points from the camera frame $\{C\}$ to $\{P_g\}$. Internally, optimization may use Euler angles for $\mathbf{R}$, but all reported rotation errors use an $SO(3)$-invariant metric.

Translation error is reported as the Euclidean distance between estimated and reference translation vectors:
\begin{equation}
\label{eq_trans_error}
e_t = \|\mathbf{t}_\text{est}-\mathbf{t}_\text{ref}\|_2,
\end{equation}
in millimeters (mm). When comparing trajectories, we report per-pose errors and trajectory summaries such as RMSE over pose indices.

Rotation error is reported as the geodesic distance on $SO(3)$:
\begin{equation}
\label{eq_rot_error}
\begin{aligned}
e_R &= \|\log(\mathbf{R}_\text{ref}^{\top}\mathbf{R}_\text{est})\|_2 \\
&= \cos^{-1}\!\left(\frac{\mathrm{trace}(\mathbf{R}_\text{ref}^{\top}\mathbf{R}_\text{est})-1}{2}\right).
\end{aligned}
\end{equation}
reported in milliradians (mrad). Euler-angle component errors are used only for visualization (e.g., per-axis drift plots) and are not treated as an invariant rotation metric.

When reporting reprojection residuals (calibration or estimator loss), we report RMS reprojection error in pixels:
\begin{equation}
\label{eq_rms_reproj}
\mathrm{RMS}_\text{px}=\sqrt{\frac{1}{N}\sum_{i=1}^N \| \mathbf{p}_i-\hat{\mathbf{p}}_i\|_2^2}
\end{equation}
where $\mathbf{p}_i$ and $\hat{\mathbf{p}}_i$ are observed and predicted pixel coordinates, respectively.

For repeated trials, we report mean $\pm$ standard deviation (SD) and, where appropriate, 95\% confidence intervals (CI) computed over independent acquisitions. When distributions are skewed or heavy-tailed, we report median and interquartile range (IQR) and use robust tests in Sec.~\ref{subsec:analysis}.

For overlap-robustness experiments (Sec.~\ref{subsubsec:overlap}), overlap ratio is defined as the fraction of valid pixels (Sec.~\ref{subsec:batching}) retained after applying the spatial partitioning mask, relative to the full valid set in that frame.

\subsection{Calibration Dataset for Hardware}
\label{subsec:calibration}
Camera and projector calibration define the geometric baseline of the proposed system. We calibrate (i) the local camera, (ii) the local projector, and (iii) the global projector using established planar-target methods and report residual statistics and dataset details for reproducibility.

We use an asymmetric circular grid target (7$\times$21) with a diameter and spacing distance of 10 mm. A total of $N_\text{pose}=16$ poses are captured for each device across a range of distances and orientations that span the operational field of view. For each pose, we detect the circle centers in the camera image domain and establish projector correspondences using phase decoding, following the projector calibration approach of Li \emph{et al.}~\cite{li2014novel}. Camera intrinsics are estimated using a standard planar-target calibration method (Zhang-style) with distortion parameters included.

For each calibrated device we estimate the intrinsic matrix and distortion parameters; extrinsics are estimated per calibration pose. For the global projector, only the intrinsics are required by the proposed estimator, but we solve the full calibration problem to obtain consistent residual statistics and discard extrinsic values during online operation because the local subsystem moves.

We report RMS reprojection error in pixels (Eq.~\ref{eq_rms_reproj}) for the camera and both projectors. Calibration statistics are summarized in Table~\ref{tab_calib_residuals}. Calibration residuals provide the empirical basis for the sensitivity/propagation analysis in Sec.~\ref{subsec:propagation} and for GT traceability statements in Sec.~\ref{subsec:setup}.

\begin{table}[!t]
\caption{Calibration dataset and residual statistics.}
\label{tab_calib_residuals}
\centering
\footnotesize
\begin{tabular}{lccc}
\toprule
Component & Poses & Points/pose & $\mathrm{RMS}_\text{px}$ \\
\midrule
Camera (local) & 16 & 147 & 0.039$\pm$0.007 \\
Projector (local) & 16 & 147 & 0.043$\pm$0.008 \\
Projector (global) & 16 & 147 & 0.044$\pm$0.008 \\
\bottomrule
\end{tabular}
\end{table}

\subsection{3D Reconstruction Accuracy and Repeatability}
\label{subsec:repeatability}
This subsection quantifies the accuracy and repeatability of the local structured-light reconstruction, which supplies the 3D geometry used by the pose estimator. We emphasize that our pose estimator does not assume a specific surface model during operation; the purpose of this subsection is solely to bound measurement error for uncertainty reporting and for sensitivity analysis (Sec.~\ref{subsec:propagation}).

We reconstruct a high-quality planar target positioned approximately fronto-parallel to the camera at a representative working distance of $d \approx 170~\mathrm{mm}$. The plane is measured across $N_\text{rep}=40$ independent acquisitions, while restarts and lighting variations are added to capture practical repeatability. For each acquisition, we apply the same phase retrieval/unwrapping pipeline (Sec.~\ref{subsec:absolute}), and we apply the measurement-validity mask described in Sec.~\ref{subsec:configuration}.

For each reconstructed point cloud $\mathbf{X}_C$, accuracy/repeatability is summarized using residuals to a best-fit plane using RANSAC on the valid region. RMS$(d)$ and its standard deviation are provided in Table \ref{tab_recon_plane}.

We quantify repeatability as the variability of the above plane metrics across independent acquisitions\cite{marrugo2020state}:
\begin{itemize}
\item within-session repeatability (back-to-back acquisitions),
\item restart repeatability (power-cycle or pipeline restart),
\end{itemize}
The intent is to bound how reconstruction variability can propagate to pose estimates; propagation is evaluated explicitly in Sec.~\ref{subsec:propagation}.

\begin{table}[!t]
\centering
\caption{3D reconstruction accuracy and repeatability on a planar target. Plane thickness is computed as the RMS distance to a best-fit plane.}
\label{tab_recon_plane}
\footnotesize
\begin{tabular}{lccc}
\toprule
Condition & Plane thickness ($\mu$m) & \# acquisitions \\
\midrule
Within-session & 7.91 $\pm$ 5.72 & $20$ \\
Restart-to-restart & 9.88 $\pm$ 7.08 & $20$ \\
\bottomrule
\end{tabular}
\end{table}

These reconstruction bounds are used in two ways: (i) to contextualize sub-millimeter pose performance claims by separating reconstruction-limited effects from estimator variability, and (ii) as the measured noise regime for the Monte-Carlo sensitivity analysis in Sec.~\ref{subsec:propagation}.

\subsection{Calibration and Measurement Error Propagation}
\label{subsec:propagation}
This subsection quantifies how calibration residuals and 3D measurement uncertainty can propagate into the estimated \emph{absolute} pose ${}^{P_g}\mathbf{T}_{C}$ returned by the proposed method. The goal is not to model every physical error source separately (e.g., thermal drift, phase noise, unwrapping mistakes), but to provide a practical sensitivity bound that links (i) measured calibration/reconstruction uncertainty to (ii) pose drift in the global frame. Estimator precision under sub-sampling is characterized separately in Sec.~\ref{subsec:characterization}.

We consider two classes of perturbations:
(i) calibration-consistent pixel perturbations that reflect finite calibration accuracy in the phase-to-pixel mapping and calibration residual statistics (Sec.~\ref{subsec:calibration}), and
(ii) \emph{measurement perturbations} affecting reconstructed 3D geometry.
Because many dominant reconstruction errors act coherently over local neighborhoods (e.g., systematic phase bias, small depth offset), we model measurement uncertainty as a spatially correlated perturbation rather than i.i.d.\ per-point noise. Specifically, for a reconstructed point $\mathbf{x}_i \in \mathbf{X}_C$, we form perturbed points
\begin{equation}
\label{eq_noise_model}
\tilde{\mathbf{x}}_i = \mathbf{x}_i + \Delta\mathbf{x}(\sigma)
\end{equation}
where $\Delta\mathbf{x}(\sigma)$ is sampled from a zero-mean Gaussian with standard deviation $\sigma$ and applied as a constant (or smoothly varying) offset over the sampled set. In addition to $\Delta\mathbf{x}(\sigma)$, we perturb the global projector pixel constraints by $\tilde{\mathbf{u}}_i^g=\mathbf{u}_i^g+\Delta\mathbf{u}(\sigma_{px})$, also as i.i.d.\ zero-mean Gaussian. \footnote{In the simplest implementation, $\Delta\mathbf{x}$ is a constant 3D offset shared by all sampled points in a trial. More structured spatial correlation models can be used, but are not required for the sensitivity results reported here.}

For each trial $k=1,\ldots,K$, we draw perturbed inputs by (a) sampling calibration perturbations $\Delta \mathbf{u}$ and (b) applying $\Delta\mathbf{x}(\sigma)$ to the reconstructed points. We then run the full estimator (Sec.~\ref{sec:model}) to convergence under the same stopping criteria and hyperparameters, producing an estimated pose ${}^{P_g}\mathbf{T}_{C}^{(k)}$. Pose drift is quantified relative to the unperturbed estimate ${}^{P_g}\mathbf{T}_{C}^{(0)}$ using the metrics in Sec.~\ref{subsec:metrics}:
\begin{equation}
\label{eq_pose_drift}
\delta_t^{(k)} = \|\mathbf{t}^{(k)}-\mathbf{t}^{(0)}\|_2,\qquad
\delta_R^{(k)} = \|\log((\mathbf{R}^{(0)})^\top \mathbf{R}^{(k)})\|_2.
\end{equation}
We report the median and 95th percentile of $(\delta_t,\delta_R)$ over $K = 20$ trials for each noise level $\sigma$.

\begin{figure}[t]
\centering
\includegraphics[width = \columnwidth]{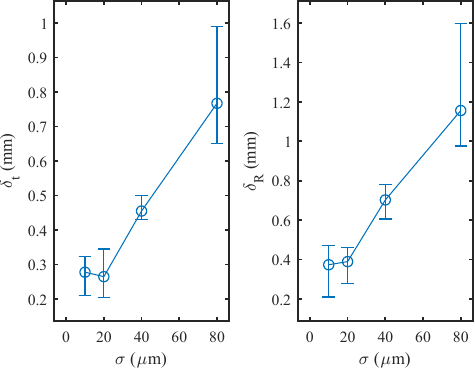}
\caption{\textbf{Pose drift under injected spatially correlated noise.} Median (IQR) drift vs perturbation magnitude $\sigma$ for (a) translation $\delta_t$ (mm) and (b) rotation $\delta_R$ (mrad) over $K=20$ trials per $\sigma$. (Sec.~\ref{subsec:propagation}).}
\label{fig_prop_curve}
\end{figure}

We first set $\sigma_0 = 10 \mu$m to the empirically measured reconstruction uncertainty from Sec.~\ref{subsec:repeatability}, and set $\sigma_{px} = 0.05 \mathrm{px}$ from the reprojection residual distribution in Sec.~\ref{subsec:calibration}. Under this measured regime, the estimator exhibits pose drift bounded by $\delta_t \leq 0.388$ mm and $\delta_R \leq 0.57$ mrad, indicating that the proposed method is not overly sensitive to the observed metrology limits of the prototype system. We additionally study failure onset for inflated perturbation magnitudes $\sigma \in \{2\sigma_0, 4\sigma_0, 8\sigma_0\}$. Details are provided in Fig.~\ref{fig_prop_curve}. Here, $\sigma_0$ is a precision (repeatability) scale used for the 3D perturbation $\Delta\mathbf{x}(\sigma)$. For brevity we report sensitivity as a function of the varying 3D perturbation level $\sigma$, while holding the calibration perturbation distribution fixed.

\subsection{Estimator Characterization}
\label{subsec:characterization}
\subsubsection{Precision and Global Sampling Invariance}
\label{subsubsec:precision}
We first characterize estimator repeatability under sub-sampling when the input 3D data and phase maps are held fixed. This isolates estimator variability due to sampling and optimization from variability due to reconstruction noise.

From a fixed reconstructed point cloud and corresponding global projector phase maps, we run the estimator for $K=400$ independent trials. Each trial draws $B=12$ batches with $n=120$ points per batch from the same validity mask (Sec.~\ref{subsec:batching}), using the default sampling scheme. The estimator is initialized with identity. We record the resulting pose estimates ${}^{P_g}\mathbf{T}_{C}^{(k)}$ and summarize their distribution.

\begin{figure}[t]
\centering
\includegraphics[width = \columnwidth]{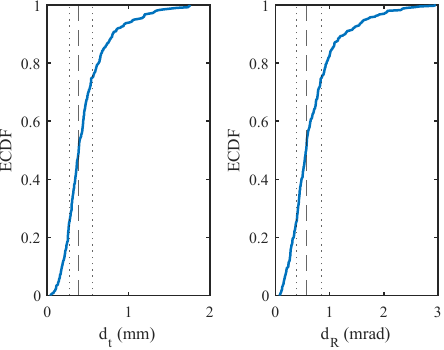}
\caption{\textbf{Sampling repeatability on fixed 3D data over $K=400$ trials.} ECDF of (a) translation repeatability $d_t$ (mm) and (b) rotation repeatability $d_R$ (mrad), defined relative to the trial-median pose. Dashed and dotted vertical lines indicate the median and IQR. (Sec.~\ref{subsubsec:precision}).}
\label{fig_sampling_repeat}
\end{figure}

Repeatability is summarized using invariant pose-dispersion metrics computed relative to the trial-median pose: translation magnitude $d_t$ and geodesic rotation angle $d_R$. Table~\ref{tab_sampling_repeat} and Fig.~\ref{fig_sampling_repeat} report summary statistics. 

\begin{table}[!t]
\centering
\caption{Sampling repeatability over $K=400$ trials on fixed 3D data. Repeatability is measured relative to the trial-median pose $\tilde{\mathbf{T}}$.}
\label{tab_sampling_repeat}
\begin{tabular}{lcc}
\toprule
Metric & $d_t=\|\mathbf{t}-\tilde{\mathbf{t}}\|_2$ (mm) &
$d_R=\angle(\tilde{\mathbf{R}}^{\top}\mathbf{R})$ (mrad) \\
\midrule
Mean $\pm$ SD & 0.37 $\pm$ 0.19 & 0.56 $\pm$ 0.31 \\
Median (IQR)  & 0.37 (0.25) & 0.55 (0.40) \\
Worst-case (max) & 0.92 & 1.75 \\
\bottomrule
\end{tabular}
\end{table}

In this fixed-data setting, estimator variability reflects sampling/optimization effects only. These results justify treating the estimator as effectively sampling-invariant at the reported operating point; additional robustness to overlap reduction and to alternative sampling schemes is evaluated next.

\subsubsection{Minimum-Overlap Robustness (Spatial Invariance)}
\label{subsubsec:overlap}
We next evaluate robustness when only a small spatial subset of valid pixels contributes constraints, modeling reduced overlap between the camera field of view and the global projector phase coverage. Starting from the same fixed reconstruction used in Sec.~\ref{subsubsec:precision}, we restrict sampling to a spatial partition of the validity mask such that only a fraction $r \in \{10\%,1\%,0.1\%,0.01\%\}$ of valid pixels can be sampled. For each overlap ratio $r$, we repeat the estimator for $K = 120$ trials and record pose distributions.

\begin{figure}[t]
\centering
\includegraphics[width = \columnwidth]{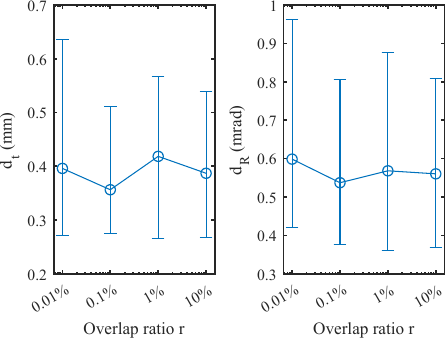}
\caption{\textbf{Minimum-overlap robustness.} Median (IQR) repeatability vs overlap ratio $r$ for (a) $d_t$ (mm) and (b) $d_R$ (mrad), over $K = 120$ trials per $r$. (Sec.~\ref{subsubsec:overlap})}
\label{fig_overlap_curve}
\end{figure}

For each overlap ratio $r$, we report invariant repeatability metrics of $d_t$ and $d_R$ relative to the trial-median pose at that $r$. Results are summarized in Fig.~\ref{fig_overlap_curve} as repeatability vs.\ overlap ratio. This experiment bounds the minimum overlap required for stable pose estimation under the measurement-validity filtering of Sec.~\ref{subsec:batching}.

\subsubsection{Sampling Strategy Comparison}
\label{subsubsec:sampling}
Finally, we evaluate invariance to sampling \emph{scheme} under matched point counts, focusing on strategies that preserve raw 3D coordinates. This distinguishes the proposed method from pipelines whose precision depends strongly on the choice of deterministic features or voxel-centroid averaging.

Using the same fixed reconstruction, we compare: (i) random sampling, (ii) uniform downsampling, and (iii) voxel-grid downsampling that retains the point nearest to each voxel center (not voxel-centroid averaging). For voxel filtering we select voxel size $s = 0.3$ mm such that the remaining point count approximately matches the random/uniform conditions. For each scheme, we run $K=120$ trials and report pose repeatability using the same invariant metrics as Sec.~\ref{subsubsec:precision}, i.e., relative to the scheme-specific trial-median pose. We summarize scheme-to-scheme differences as relative increases in median(IQR). Results are shown in Table~\ref{tab_scheme_compare}.

\begin{table}[!t]
\centering
\caption{Sampling scheme comparison under matched point counts. Reported as median (IQR) over $K=120$ trials per scheme.}
\label{tab_scheme_compare}
\begin{tabular}{lccc}
\toprule
Scheme & Points & $d_t$ (mm) & $d_R$ (mrad) \\
\midrule
Random & 501350 & 0.37 (0.26) & 0.52 (0.37) \\
Uniform & 501350 & 0.30 (0.23) & 0.49 (0.38) \\
Voxel & 492988 & 0.39 (0.27) & 0.56 (0.41) \\
\bottomrule
\end{tabular}
\end{table}

We do not include voxel-centroid averaging in this comparison because it modifies the underlying 3D coordinates and therefore changes the measurement model rather than the sampling scheme. Such coordinate-modifying filters are known to introduce bias when the downstream estimator assumes higher depth accuracy than the voxel size.

\subsubsection{Hyperparameters}
\label{subsubsec:hyperparameters}
We report the exact optimization settings used for all experiments to ensure reproducibility, provided in Table.\ref{tab:opt_defaults}.

\begin{table}[!t]
\centering
\caption{Optimization defaults used throughout the paper and tested tolerance ranges.}
\label{tab:opt_defaults}
\begin{tabular}{lcc}
\toprule
Setting & Default & Tested range \\
\midrule
Batch size $B$ & 12 & 5--400 \\
Points per batch $n$ & 120 & 1--400 \\
Max iters $T$ & 2000 & 20--200000 \\
Adam $\alpha_0$ & 1e-2 & 5e-4--1e-1 \\
Adam $(\beta_1,\beta_2,\epsilon)$ & (0.9, 0.999, 1e-8) & fixed \\
Stage-2 weight $\lambda$ & 0.01 & 0.001--1 \\
Early stop threshold & 1e-7 & 1e-11--1e-4 \\
\bottomrule
\end{tabular}
\end{table}

\section{Experiments}
\label{sec:experiments}
This section evaluates the proposed projector-referenced pose estimation framework in two settings where ICP is known to show noticeable error. First, we assess absolute pose estimation and loop-closure behavior on a featureless planar surface, where conventional ICP is ill-posed due to the absence of geometric features. Second, we evaluate error accumulation and sensitivity to downsampling in a sequential registration scenario, comparing ICP to ICP+Model under controlled voxel-grid filtering. Unless otherwise stated, metrics and reporting conventions (translation magnitude, SO(3) rotation error, uncertainty intervals) follow Sec.~\ref{subsec:metrics}.

\subsection{Experimental Setup, Acquisition, and Ground Truth}
\label{subsec:setup}
A two-projector/one-camera structured-light system was constructed (Fig.~\ref{setup_img}). The setup uses two digital light processing (DLP) projectors (Texas Instruments LightCrafter 4500; 912×1140) and a monochrome camera (FLIR GS3-U3-32S4M; 2448×2048). The global projector remains fixed and defines the global reporting frame used throughout the manuscript, while the local structured-light subsystem (camera + local projector) is moved within the global projector’s field of view.

The experiments are performed in a quasi-static manner to isolate estimator behavior and to support the ground-truth construction. Acquisition was time-multiplexed to prevent pattern interference. The camera and projectors are synchronized using an external trigger, where the exposure time is set to 8.3ms (120FPS). Local structured-light reconstruction uses 23 frames (18-step phase shift + 5 Gray code), and the global projector constraint uses 22 frames (6-step phase shift + 5 Gray code per axis), for a total of 45 frames per pose update. This implies an acquisition-limited update rate of approximately 2.7 pose updates/s under the experimental schedule. Estimator-only runtime was reported separately in Table~\ref{tab:gt_timing} to show that computation is not the bottleneck; throughput is dominated by the programmed pattern schedule and camera frame rate rather than by the solver. 

Additional details on phase retrieval and reconstruction are provided in Sec.~\ref{subsec:absolute}. Camera/projector calibration procedures, dataset sizes, and calibration residuals are reported additionally in Sec.~\ref{subsec:calibration}. Reconstruction accuracy and repeatability are characterized in Sec.~\ref{subsec:repeatability}. These sections define the metrological baseline used for uncertainty reporting and for the propagation/sensitivity analysis in Sec.~\ref{subsec:propagation}.

ICP uses a point-to-plane objective with normal estimation with the following fixed settings: maximum iterations 100, convergence criteria 5e-3, and normal estimation neighborhood 20. To ensure comparability across conditions, the downsampling protocol used to stress ICP (voxel-grid selection of nearest-to-voxel-center points) is applied consistently wherever voxel filtering is reported. 

Ground truth (GT) differs across the two experiments and is summarized in Table~\ref{tab:gt_timing} along with conservative expanded uncertainty bounds.
For the plane experiment, translation GT is derived in the global projector frame and rotation is bounded by a pixel-level alignment uncertainty at the stated working distance of $d \approx 170~\mathrm{mm}$. Specifically, assuming a conservative alignment uncertainty of $\sigma_{px}^{GT}$ pixels in the global projector image, we bound angular uncertainty by $\sigma_R^{GT}\approx \sigma^{GT}_{px}/f_g$ (small-angle) and report the translation-equivalent bound $d\,\sigma_R^{GT} \approx 0.014~\mathrm{mm}$, where $f_g$ is the projector focal length in pixel units.
For the trajectory experiment, GT at discrete placements is defined by a repeatable marker/fixture procedure and manual alignment; translation uncertainty is dominated by the alignment step. When measured errors approach the GT expanded uncertainty, results are interpreted as GT-limited rather than as evidence of sub-GT performance\cite{jcgm100_2008}.

\begin{table*}[!t]
\centering
\caption{Ground truth (GT) traceability and timing summary for Sec.~\ref{subsec:setup}. Estimator-only timing excludes acquisition and reconstruction. Success rate is defined as batch hypotheses convergence rate (individual pose) and frame success rate (result).}
\label{tab:gt_timing}
\begin{tabular}{p{0.48\textwidth} p{0.48\textwidth}}
\toprule
\textbf{GT traceability (used for $e_t,e_R$)} &
\textbf{Timing summary }\\
\midrule
\emph{Plane registration:} GT pose source = Derived from calibrated global-projector frame $\{P_g\}$ using projector residual statistics (Sec.~\ref{subsec:calibration}), \newline
Expanded uncertainty (95\%): $\sigma_t^{GT}$ = 0.01 mm, $\sigma_R^{GT}=0.0833$ mrad  \newline
\emph{ICP correction:} GT trajectory source = Discrete pose locations established by physical markers; marker centers estimated in the calibrated global projector frame $\{P_g\}$; pose evaluated at marked placements, alignment provided using calipers and visual edge. Translation uncertainty dominated by caliper resolution (0.05 mm) and alignment repeatability. 
&
Acquisition schedule (per pose/frame): local = 18-step + Graycode (5 images), global = 6-step per direction + Graycode (5 images) \newline
Achieved acquisition rate: 2.5--3 recon/s (limited by camera and sequence) \newline
Reconstruction+preprocess throughput (offline): $\sim$60 recon/s \newline
Estimator-only throughput: median $\sim$62 Hz, p95 $\sim$21 Hz across 400 samples \newline
Success rate under stopping criteria: 83\% (individual pose), 97\% (result)
\\
\bottomrule
\end{tabular}
\end{table*}

\subsection{Featureless Surface Registration and Loop Closure}
\label{subsec:featureless}
This experiment evaluates whether the proposed method can estimate pose and register point clouds on a surface that is geometrically featureless, where ICP is fundamentally underconstrained (translation along the plane and rotation about the plane normal).

A planar target with maximum out-of-flatness below 10 $\mu$m was used. A rectangular trajectory with 6 discrete sensor placements was defined by pre-marking expected camera fields of view such that the merged reconstruction forms a known planar region. Intermediate poses include partial overlap conditions between the local system and global projector field of view to evaluate robustness under reduced overlap.

For each placement, the local structured-light subsystem acquires a reconstruction, and the proposed estimator computes the sensor pose directly with respect to the fixed global projector reference. Each reconstructed point cloud is transformed into the global reference and merged without requiring sequential chaining, so loop closure is assessed by comparing the final pose against the initial pose and by evaluating the geometric consistency of the merged surface.

\begin{figure*}[t]
\centering
\includegraphics[width=\linewidth]{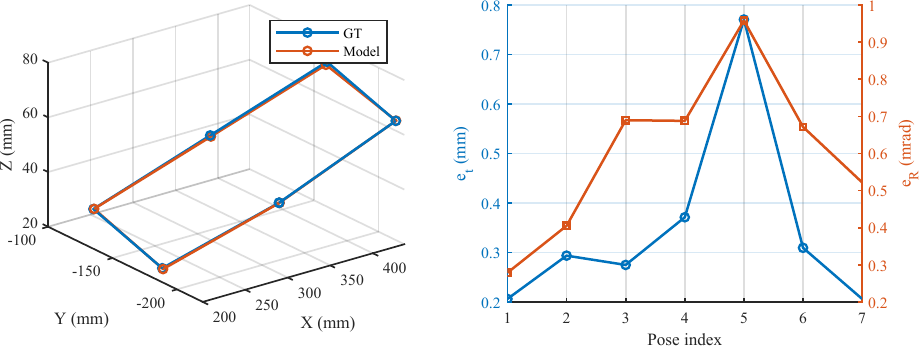}
\caption{\textbf{Featureless plane registration.} (a) Estimated trajectory vs GT in the global reference frame. (b) Per-pose pose error relative to GT: translation $e_t$ (mm, left axis) and rotation $e_R$ (mrad, right axis). (Sec.~\ref{subsec:featureless})}
\label{fig_plane_trajectory}
\end{figure*}

We report pose error vs GT for each placement, including max and RMS values over the trajectory in Fig.~\ref{fig_plane_trajectory}, along with the 3D trajectories comparing the proposed model with the constructed GT in the global reference frame. 

We additionally evaluate post-registration alignment quality on the overlapping region using correspondence-free geometric metrics:
(i) symmetric bidirectional cloud-to-cloud distance (mean nearest-neighbor distance in both directions)\cite{girardeau2005change}, and
(ii) trimmed Hausdorff distance (95th percentile of nearest-neighbor distances)\cite{chui2003new}.
Across placements, the symmetric cloud distance was median 0.12 mm (IQR 0.05) and the trimmed Hausdorff distance was median 0.16 mm (IQR 0.08),
supporting sub-millimeter alignment quality consistent with the pose-error results in Fig.~\ref{fig_plane_trajectory}.

On this featureless surface, ICP becomes ill-conditioned and yields inconsistent translation estimates. In contrast, the proposed method estimates the trajectory with a maximum translation deviation of 0.86 mm, and the merged surface dimensions deviate from the predetermined FOV by 0.6\%.

\subsection{Error Accumulation Reduction in Sequential Registration}
\label{subsec:sequential}
This experiment evaluates whether integrating the proposed estimator with ICP reduces (i) drift/error accumulation in a pose chain and (ii) sensitivity to downsampling settings, which commonly introduce precision artifacts in conventional point-to-plane ICP.

A rigid object (statue) was reconstructed at 8 discrete sensor placements (7 poses excluding the initial pose). Each raw point cloud contains roughly 2.0M points. The sensor is repositioned between poses while maintaining visibility within the global projector field of view. 

To emulate practical real-time constraints and induce ICP sensitivity, each point cloud is downsampled prior to registration. We vary the voxel-grid size $s$ over $[0.3,3.0]$~mm using $N_v=20$ discrete settings. For voxel-grid downsampling, we retain the point nearest each voxel center (i.e., no centroid averaging) so that raw 3D coordinates are not modified; this isolates the effect of reduced point density from coordinate quantization. For each voxel size $s$, ICP registers pose $i$ to pose $i-1$ using the configuration described in Sec.~\ref{subsec:setup}. Relative transforms are composed to obtain the full trajectory. This produces one deterministic trajectory per voxel size.

For each registration step and voxel size, ICP provides an initial estimate which initializes the proposed estimator. The estimator then refines the pose using projector-referenced constraints via batched sampling and outlier rejection. The corrected transform is composed to produce the trajectory. 

\begin{figure*}[t]
\centering
\includegraphics[width=\linewidth]{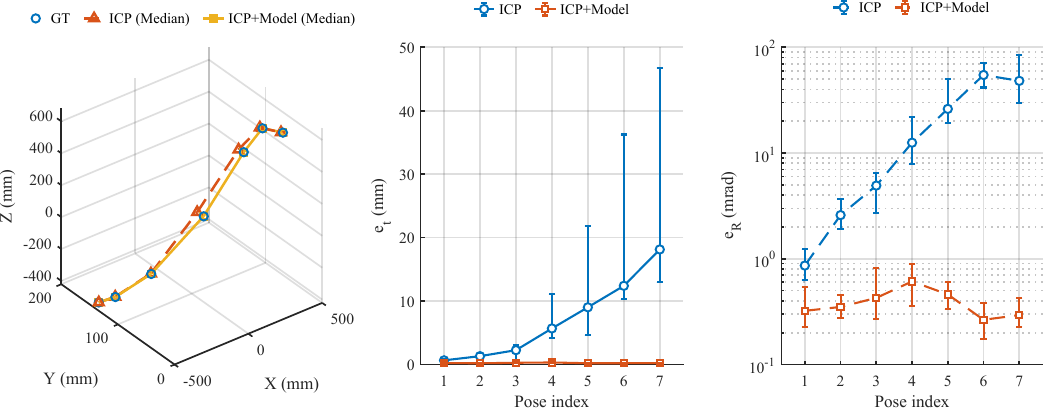}
\caption{\textbf{Error accumulation reduction under downsampling.} Median trajectories for ICP and ICP+Model. GT markers are shown for reference. ICP+Model closely overlaps GT; separation is quantified in the right panel. (b) Per-pose translation error $e_t$ (mm) relative to GT (median with IQR over voxel sizes). (c) Per-pose rotation error $e_R$ (mrad) relative to GT (median with IQR over voxel sizes). The y-axis is logarithmic to accommodate the large dynamic range between ICP and ICP+Model. (Sec.~\ref{subsec:sequential}).}
\label{fig_statue_trajectory}
\end{figure*}

Because ICP produces relative transforms, the primary evaluation uses relative pose error along the chain compared to GT, reported as translation magnitude and SO(3) rotation error per pose, and summarized as trajectory RMSE across poses. To visualize sensitivity to voxel size, we report the median trajectory with an error band across voxel sizes (e.g., 5–95 percent range) and per-pose sensitivity statistics (e.g., IQR across voxel sizes). Fig.~\ref{fig_statue_trajectory} illustrates drift/error accumulation and precision artifacts for ICP and ICP+Model across a sweep of downsampling settings.

ICP exhibits increasing drift and sensitivity to voxel size, reaching 18.1 mm translation error and 48.3 mrad rotation error by pose 7, and showing substantial spread across voxel sizes. Adding the proposed estimator substantially reduces voxel-induced variability and reduces drift, with translation error remaining below 0.23 mm and rotation error below 0.29 mrad across the full chain.

\subsection{Statistical Analysis}
\label{subsec:analysis}
To confirm that improvements in accuracy and robustness are statistically significant, we perform paired tests comparing ICP and ICP+Model under matched conditions. The paired unit is the voxel size condition for accuracy metrics ($n = 20$) and the pose index for voxel-sensitivity metrics ($n = 7$).

For each voxel size $s$ and method $m$, define per-pose errors $e_t(i, s, m)$ and $e_R(i, s, m)$ against GT (Sec.~\ref{subsec:metrics}). We summarize each voxel condition with a scalar trajectory score,

\begin{equation}
\label{scalar_trajectory_score}
\begin{aligned}
E_T &= \sqrt{\frac{1}{N}\sum_{i=1}^{N} \left\| \mathbf{t}_{\text{est}}(i) - \mathbf{t}_{\text{GT}}(i) \right\|_2^2},\\
E_R &= \sqrt{\frac{1}{N}\sum_{i=1}^{N} \left( e_R\!\left(\mathbf{R}_{\text{est}}(i),\mathbf{R}_{\text{GT}}(i)\right) \right)^2 }.
\end{aligned}
\end{equation}

and test whether paired differences $E(s, \text{ICP}) - E(s, \text{ICP+Model})$ are greater than zero. We report median improvement, 95 percent confidence intervals (bootstrap over paired units), and p-values using a paired nonparametric test (Wilcoxon signed-rank). Multiple comparisons across translation/rotation metrics are handled using statements that results are unchanged under correction.

Robustness test (voxel-sensitivity across pose indices). To quantify sensitivity to voxel size (precision artifacts), we compute per-pose spread across voxel sizes, e.g.,
\begin{equation}
\label{per_pose_spread}
\begin{aligned}
\sigma_T(i) &= \operatorname{Std}\!\left(\left\{ \left\| \mathbf{t}(i)-\bar{\mathbf{t}}(i) \right\|_2 \right\}\right),\\
\sigma_R(i) &= \operatorname{Std}\!\left(\left\{ e_R\!\left(\mathbf{R}(i),\bar{\mathbf{R}}(i)\right) \right\}\right).
\end{aligned}
\end{equation}

and compare $S(i, \text{ICP})$ and $S(i, \text{ICP+Model}$ across pose indices. Results are summarized as median reduction factors with confidence intervals.

Statistical results are summarized in Table ~\ref{tab:stats}, including test type, sample size, median improvement, CI, and p-value. Where errors approach GT expanded uncertainty, metrics are interpreted as GT-limited per Sec.~\ref{subsec:setup}.

\begin{table}[!t]
\centering
\caption{Paired Wilcoxon signed-rank tests for ICP vs ICP+Model. $\Delta$ denotes paired improvement (ICP$-$ICP+Model); positive values favor ICP+Model. 95\% CIs are bootstrap percentile over paired units.}
\label{tab:stats}
\begin{tabular}{lccccl}
\toprule
Metric & Unit & $n$ & Median $\Delta$ & 95\% CI & $p$ \\
\midrule
$E_T$ & mm   & 20 & 9.22  & 7.27, 15.80 & $4.8\times10^{-5}$ \\
$E_R$ & mrad & 20 & 29.28 & 20.95, 42.36 & $4.8\times10^{-5}$ \\
$\sigma_T$ & mm   & 7 & 6.94  & 0.80, 33.52 & 0.156 \\
$\sigma_R$ & mrad & 7 & 9.40  & 1.25, 29.97 & 0.156 \\
\bottomrule
\end{tabular}
\end{table}

\section{Conclusion}
\label{sec:conclusion}
We introduced an auxiliary-hardware pose estimator for dense DFP mapping that uses a fixed, intrinsically calibrated global projector as a reference. Rather than relying on point-to-point correspondences between reconstructed clouds, the estimator constrains the moving DFP system pose through phase-derived pixel observations and a reprojection objective, stabilized by batched subsampling and consensus-based rejection. Across the reported characterization and application experiments, the method maintained sub-millimeter pose accuracy with high repeatability under aggressive coordinate-preserving subsampling, enabled registration on homogeneous surfaces and low-overlap views where correspondence-based ICP is unreliable, and reduced trajectory error accumulation when used to correct ICP-based registration, with statistical significance reported in Sec.~\ref{subsec:analysis}.

The proposed approach introduces clear trade-offs. First, it requires auxiliary hardware and a fixed global projector reference, which constrains deployability and may require re-calibration or verification when the global projector or mounting changes. Second, the global and local measurements are time-multiplexed to avoid pattern interference; therefore, the achievable end-to-end update rate is limited by the additional pattern sequence and exposure constraints, and simultaneous measurement is not assumed. Third, failure can occur when the global projector constraints are degraded, e.g., under saturation, specular reflections, severe occlusion, fringe/phase unwrapping errors, extreme poses that reduce usable overlap, or insufficient valid pixels after masking.

From a practical standpoint, the estimator exposes a direct quality indicator: the reprojection residual in the global projector pixel domain. This residual can be monitored online to detect constraint degradation and trigger reacquisition, re-initialization, or recalibration checks, which is particularly useful in long capture sessions where drift or gradual misalignment may occur. The sensitivity analysis in Sec.~\ref{subsec:characterization} provides a quantitative stability regime for expected calibration and measurement uncertainty, and a stress regime illustrating degradation under larger perturbations, which can guide acceptance testing and deployment envelopes. It additionally motivates follow-on isolation of dominant contributions, such as systematic phase bias.

Future work will focus on reducing acquisition overhead and improving deployability. A primary direction is to reduce the number of global-projector patterns (or replace them with a lightweight pose-constraint pattern) while maintaining pose repeatability, and to investigate partial or simultaneous projection strategies that remove strict time multiplexing. A second direction is to extend the framework beyond a single fixed projector, e.g., multiple global references or wider-coverage illumination, and to incorporate automatic health checks and periodic recalibration protocols. A matched comparison against alternative robust dense registration baselines under identical subsampling is a complementary direction for future study.

\section*{Acknowledgements}
This research was supported by the Technology Innovation Program(Project Name: Development of AI autonomous continuous production system technology for gas turbine blade maintenance and regeneration for power generation, Project Number: RS-2025-25447257, Contribution Rate: 50\%) funded By the Ministry of Trade, Industry and Resources(MOTIR, Korea), the Culture, Sports and Tourism R\&D Program through the Korea Creative Content Agency grant funded by the Ministry of Culture, Sports and Tourism in 2024 (Project Name: Global Talent for Generative AI Copyright Infringement and Copyright Theft, Project Number: RS-2024-00398413, Contribution Rate: 40\%), and the National Research Foundation of Korea(NRF) grant funded by the Korea government(MSIT) (Project Number: RS-2025-16072782, Contribution Rate: 10\%).

\vspace{150px}
\bibliography{ref}
\vspace{200px}
\bibliographystyle{IEEEtran}

\begin{IEEEbiography}
[{\includegraphics[width=1in,clip,keepaspectratio]{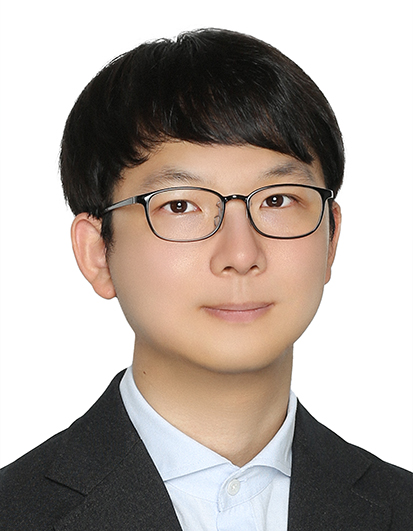}}]
{Sehoon Tak} received his B.S. degree in Mechanical Engineering from Korea Advanced Institute of Science and Technology(KAIST), Daejeon, South Korea. He is currently pursuing a Ph.D. degree at Yonsei University, Seoul, South Korea, in mechanical engineering. His current research interests include 3D optical sensing, robotics, and 3D reconstruction.
\end{IEEEbiography}
\begin{IEEEbiography}
[{\includegraphics[width=1in,clip,keepaspectratio]{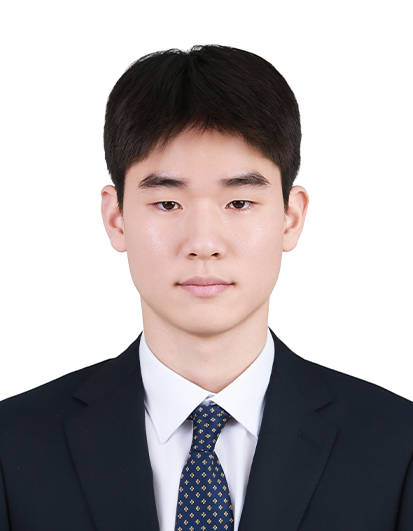}}]
{Keunhee Cho} received his B.S. degree in Mechanical Engineering from Yonsei University, Seoul, South Korea, in 2025, where he is currently pursuing an M.S. degree in the same field. His current research interests include 3D optical sensing, robotics, and virtual reality.
\end{IEEEbiography}
\begin{IEEEbiography}[{\includegraphics[width=1in,clip,keepaspectratio]{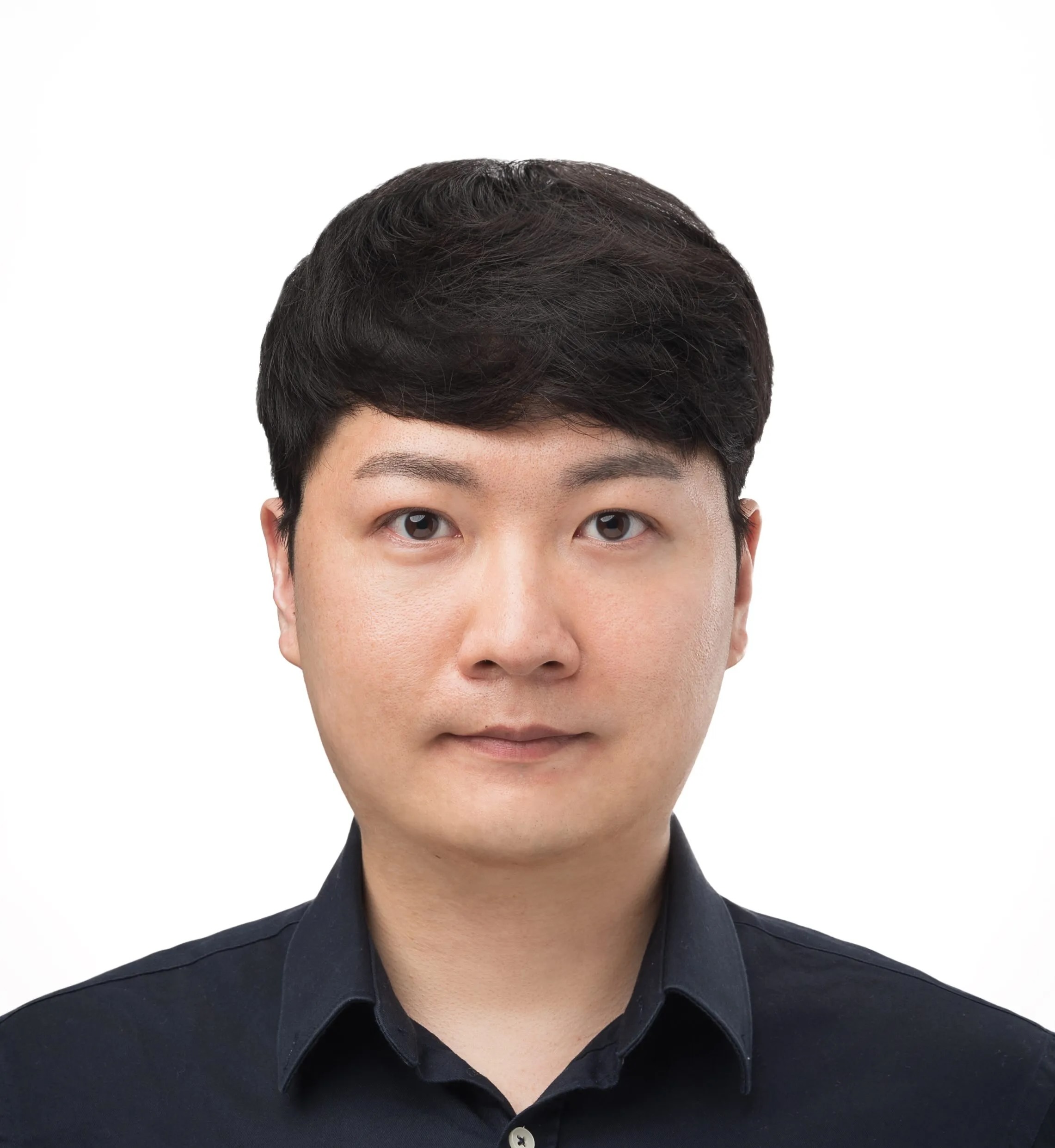}}]{Sangpil Kim} is an assistant professor in the Department of Computer Science and Engineering at Korea University. He received his Ph.D. in Electrical and Computer Engineering from Purdue University and earned his B.S. degree in Computer Science from Korea University, South Korea. His current research focuses on the intersection of computer vision and deep learning, with an emphasis on applications of multi-modal fusion for developing generative models.
\end{IEEEbiography}
\begin{IEEEbiography}
[{\includegraphics[width=1in,clip,keepaspectratio]{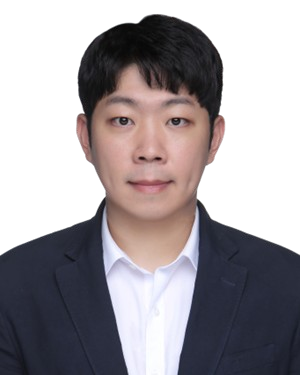}}]{Jae-Sang Hyun}
   is an assistant professor in the Department of Mechanical Engineering, Yonsei University. He worked at ORBBEC 3D, Troy, MI, USA as a research scientist for the development of 3D cameras. He received his Ph.D. in mechanical engineering from Purdue University and B.S. degree in mechanical engineering from Yonsei University, Seoul, South Korea. His major research areas include 3D optical sensing, 3D reconstruction, and 3D sensor development.
\end{IEEEbiography}

\end{document}